\documentclass[11pt]{article}
\usepackage{acl2016}
\usepackage{times}
\usepackage{url}
\usepackage{latexsym}

\usepackage{url}
\usepackage{graphicx}
\usepackage{multirow}
\usepackage{latexsym}
\usepackage{amsmath}
\usepackage[pass]{geometry}
\usepackage{booktabs} 

\newenvironment{myquote}{\list{}{\leftmargin=0.1in\rightmargin=0.1in}\item[]}{\endlist}

\aclfinalcopy 
\def\aclpaperid{518} 

\title{Verbs Taking Clausal and Non-Finite Arguments as Signals of Modality -- Revisiting the Issue of Meaning Grounded in Syntax}

\author{Judith Eckle-Kohler \\
Research Training Group AIPHES and UKP Lab\\
Computer Science Department, Technische Universit\"at Darmstadt \\
  {\tt www.aiphes.tu-darmstadt.de, www.ukp.tu-darmstadt.de}
}

\date{}
\begin{document}

\maketitle

\begin{abstract}
We revisit Levin's theory about the correspondence of verb meaning and syntax and
infer semantic classes from  a large syntactic classification of more than 600 German verbs
 taking clausal and non-finite arguments. 
Grasping the  meaning components of Levin-classes is known to be hard.
We address this challenge by setting up a multi-perspective semantic characterization of  the inferred classes.
To this end, we link
the inferred classes and their English translation to independently constructed semantic classes in three different lexicons -- the German wordnet GermaNet, VerbNet and FrameNet -- and perform a detailed analysis and evaluation of the resulting
German--English classification (available at \url{www.ukp.tu-darmstadt.de/
modality-verbclasses/}). 

\end{abstract}

\section{Introduction}
\label{intro}
Verbs taking clausal and non-finite arguments add a further meaning component to
their embedded argument. For example, 
the embedded argument is realized as \emph{that}-clause in (1) and (2),
but \emph{understand} in (1) marks it as factual and \emph{hope} in (2) as uncertain.
The verb \emph{pretend} in (3) realizes its embedded argument as non-finite construction and marks it as non-factual.
\vspace{-0.3cm}
\begin{myquote}
(1) \emph{He \underline{understands} \textbf{that his computer has a hardware problem}.}

(2) \emph{She \underline{hopes} \textbf{that her experience will help others}.}

(3) \emph{He \underline{pretends} \textbf{to take notes on his laptop}, but really is updating his Facebook profile.}
\end{myquote}
The entities expressed by embedded clausal and non-finite arguments are also called ``abstract object'' (AO) in the rest of this paper
(following \newcite{asher93}); we will  use the linguistic term ``modality''  \cite{valentine2011} to subsume the meanings (such as factuality, non-factuality and uncertainty) denoted by AO-selecting verbs.

As AO-selecting verbs can change the meaning of a text in important
ways, text understanding systems should be sensitive to them. 
In particular,
classifications of AO-selecting verbs according to semantic criteria 
are important knowledge sources for a wide range of NLP applications, such as
event tagging \cite{Sauri:2005}, commited belief tagging \cite{prabhakaran-rambow-diab:2010}, reported speech tagging \cite{KRESTEL08.718}, the detection of uncertainty  \cite{TUD-CS-2012-0035}
and future-oriented content \cite{TUD-CS-20081509},
textual entailment \cite{Saurf:2007,lotan-stern-dagan:2013:NAACL-HLT}, or 
determining the degree of factuality of a given text \cite{Sauri:2012,deMarneffe:2012}.
Accordingly, various semantic classifications of AO-selecting verbs have been developed, e.g., 
\cite{kiparsky70,karttunen1971implicative,karttunen:2012}, 
some of them explicitly in the context of
NLP  \cite{nairn2006computing,sauri2008}.

However, these classifications 
are constructed manually and often quite limited in coverage. Consequently, extending  or adapting them to specific domains or other languages is a major issue.

We propose to address this issue by exploiting the relationship between the syntactic behavior of verbs and their meaning following Levin's theory \cite{levin93}.
This has not been done yet for verbs signaling modality, as far as we are aware.
For the particular category of AO-selecting verbs, Levin's theory allows constructing verb classifications in a purely syntax-driven way, i.e. inducing semantic classes from syntactically defined classes, and thus possibly also extending given classes using large corpora.\footnote{Abstract objects already characterize the possible semantic roles to a certain extent.}

While the appeal of Levin's hypotheses is clear, we are aware of a major difficulty, making our approach a challenging research problem:  it is very hard to grasp the precise meaning components which are to be associated with a syntactic ``Levin'' class. At the same time, it is vital to have a good semantic characterization of the meaning components in order to apply  such classes to NLP tasks in an informed way.

We address these issues and  make the following contributions:
(i) We consider a purely syntactic classification of more than 600 German AO-selecting verbs and induce semantic classes
based on findings from formal semantics about correspondences between verb syntax and meaning. This yields an initial description of the meaning components associated with the classes, along with a tentative class name.
(ii) In a second step, we refine and extend the semantic characterization of the verb classes by translating it to English and  linking it
to existing semantic classes in lexical resources at the word sense level:
we consider the coarse semantic fields in the German wordnet GermaNet \cite{Kunze02}, 
the verb classes in the English lexicon VerbNet \cite{kipper2008}, and the semantic frames in the English lexicon FrameNet \cite{Baker1998}.
As a result, we obtain a detailed semantic characterization of the verb classes, as well as insights into the validity of Levin's theory across the related languages German and English.
(iii) We also perform a task-oriented evaluation
of the verb classes in textual entailment recognition, making use of insights from the previous two steps.
The results
suggest that the verb classes might be a promising resource for this task,
for German and for English.

\section{Related Work}
This section 
summarizes related work about the correspondence between verb meaning and syntax and discusses related work on modality  in NLP.

{\bf Syntactic Reflections of Verb Meaning}
Semantic verb classifications that are grounded in lexical-syntactic properties of verbs are particularly appealing, because they
can automatically be recovered in corpora based on syntactic features.
The most well known verb classification based on correspondences between
 verb syntax and verb meaning is Levin's classification  \cite{levin93}.
According to \newcite{Levin2015}, verbs that share common syntactic argument alternation patterns also have particular
meaning components in common, thus they can be grouped into a semantic verb class.
For example, verbs participating in the dative alternation\footnote{These verbs can realize an argument syntactically either as 
noun phrase or as prepositional phrase with \emph{to}.}
can be grouped
into a semantic class of verbs sharing the particular meaning component  ``change of possession'', thus this shared meaning component characterizes the semantic class.
Recent work on verb semantics provides additional evidence for this correspondence of verb syntax and meaning: 
\newcite{HartshorneBP14} report that
the syntactic behavior of some verbs can be predicted based on their meaning.

VerbNet is a broad-coverage verb lexicon organized in verb classes based on Levin-style syntactic alternations:
verbs with common subcategorization frames and syntactic alternation behavior that also share common semantic roles are grouped
 into  VerbNet classes. 
VerbNet not only includes the verbs from the original verb classification by Levin, but also more than 50 additional 
verb classes \cite{kipper:lrec2006} automatically acquired from corpora \cite{Korhonen:2004}. These classes contain many 
AO-selecting verbs that were not covered by Levin's classification.
However, VerbNet does not provide information about the modal meaning of AO-selecting verbs and does not reflect fine-grained distinctions between various kinds of modality.

There is also some criticism in previous work regarding the validity of Levin's approach.
\newcite{BakerRuppenhoferBLS2002} and \newcite{schnorbusch} both discuss various issues with Levin's original classification, 
in particular the difficulty to grasp the meaning components, 
which are to be associated with a Levin class.

While approaches to exploit the syntactic behavior of verbs for the automatic acquisition of semantic verb classes from corpora have been developed in the past, they were used to recover only small verb classifications: \newcite{SchulteImWalde06}'s work considered a semantically balanced set of 168 German verbs, \newcite{Merlo2001} used 60 English verbs from three particular semantic classes.

In contrast to previous work, we consider a large set of more than 600 German AO-selecting verbs and focus on their modal meaning (i.e., expressing factuality or uncertainty).

{\bf Related Work on Modality in NLP} 
Previous work in NLP on the automatic (and manual) annotation of modality has often tailored the concept of modality to particular
applications.
\newcite{TUD-CS-2012-0035} introduce a taxonomy of different kinds of modality expressing uncertainty,
such as  deontic, bouletic, abilitative modality,
 and use it for detecting uncertainty in an
Information Extraction setting. Their uncertainty cues also include verbs.

\newcite{Sauri:2012} use discrete values in a modality continuum ranging from
uncertain to absolutely certain in order to automatically determine the factuality of events mentioned in text. Their automatic approach is based on the FactBank corpus \cite{factbank}, a corpus of newswire data with manually annotated event mentions.
For the factuality annotation of the event mentions, the 
human annotators were instructed to primarily base their decision on lexical cues.
For example, they used verbs of belief and opinion, perception verbs, or verbs expressing proof.

\newcite{Nissim2013} introduce an annotation scheme for the cross-linguistic annotation of modality in corpora. 
Their annotation scheme defines two dimensions which are to be annotated (called layers): factuality 
 (characterizing the embedded proposition or concept) and speaker's attitude (characterizing the embedding predicate). 
 Their annotation scheme starts from a fixed set of modal meanings 
  and aims at finding previously unknown triggers of modality.
However, some modal meanings are not 
distinguished, in particular those involving future-orientation.
A classification approach grounded in syntax -- as in our work -- can be considered as complementary: it starts from
the syntactic analysis  of a large set of trigger words, and induces a broad range of modal meanings
based on correspondences between verbs syntax and meaning.

Our semantic classification for
AO-selecting verbs covers a \emph{wide range} of different kinds of modality in text, thus considerably extending previous work.

\section{Inferring Semantic Verb Classes}
In this section, we  infer semantic verb classes from the syntactic 
alternation behavior of a large dataset of German AO-selecting verbs.
The research hypotheses underlying our method can be summarized as follows:
There are correspondences between verb syntax and meaning: certain syntactic alternations correspond to
particular meaning components \cite{Levin2015}.

\subsection{German Subcategorization Lexicon}
We consider a set of 637 AO-selecting verbs given in \cite{eckle99}. These verbs
are a subset of a subcategorization lexicon (i.e., pairs of lemma and subcategorization frame)
 that has automatically been extracted from large 
newspaper corpora
using a shallow regular expression grammar covering more than 240 subcategorization frames (short: subcat frames). All the subcat frames extracted for a given verb were manually checked and only the correct ones were included in the final lexicon, because high quality lexical information was crucial in the target application Lexical Functional Grammar parsing.\footnote{Today, this lexicon is part of the larger resource ``IMSLex German Lexicon'' \cite{fitschen2004}.} 

\newcite{eckle99} specified
the alternation behavior of each AO-selecting verb regarding different types of clausal and non-finite arguments, yielding a \emph{syntactic signature} for each verb (e.g., 111101 for the verb 
\emph{einsehen (realize)} using the encoding in Table \ref{synPrimitives}, top to bottom corresponding to left to right).\footnote{The automatically extracted subcategorization lexicon also contains adjectives and nouns taking clausal or infinitival arguments. However, many of the 1191 nouns and 666 adjectives are derived from verbs, which makes them the central word class.}
For this, each verb was inspected regarding its ability to
take any of the considered clausal and non-finite constructions as argument
-- either on the basis of the automatically acquired subcat frames or
by making use of linguistic introspection. Linguistic
 introspection is necessary to reliably identify non-possible  argument types,
since missing subcat frames  that were not extracted automatically are not sufficient as evidence.

\begin{table}[h]
\footnotesize
\begin{center}
\begin{tabular}{@{}p{3.5cm}ll}
\hline\noalign{\smallskip}
\bf Argument Type   & Y$/$N & \bf Example \\ 
\noalign{\smallskip}\hline\noalign{\smallskip}
\emph{da{\ss}(that)}-clause  & 1$/$0 & \emph{sehen (see)}\\
\emph{zu(to)}-infinitive, present & 1$/$0 & \emph{versuchen (try)} \\
\emph{zu(to)}-infinitive, past & 1$/$0 & \emph{bereuen (regret)}\\
\emph{wh}-clause  & 1$/$0 & \emph{einsehen (realize)} \\
\emph{ob(whether/if)}-clause & 1$/$0 &  \emph{fragen (ask)} \\
declarative clause & 1$/$0 & \emph{schreien (shout)} \\
\noalign{\smallskip}\hline
\end{tabular}
\end{center}
\caption{Clausal and infinitival arguments distinguished in the syntactic classification;
possibility of each type is encoded as 1 (possible) or 0 (not possible).}
\label{synPrimitives}
\end{table}

Although there are 64 possible syntactic signatures according to basic combinatorics, in the data only 46 signatures were found,
which group the verbs into 46 classes.
While \newcite{eckle99} points out a few semantic characteristics of these classes, 
most of them lack a semantic characterization. Our goal is to address this gap and to
infer shared meaning components for all the classes. For this, we use linguistic research findings as described in the next section.

\subsection{Findings from Formal Semantics} \label{hypo}
\begin{table*}[t]
\centering
\footnotesize
\begin{tabular}{p{1cm}p{3.5cm}|p{3.7cm}|p{6.1cm}}
\hline\noalign{\smallskip}
\bf signature & \bf \#verbs -- examples & \bf meaning components  &  \bf semantic characterization (\#linked verbs) \\
 \noalign{\smallskip}\hline\noalign{\smallskip}
 
\textbf{010} {-}{-}{-} & 36 (6\%) -- \emph{wagen (dare), z\"{o}gern (hesitate), weigern (refuse)} & {\bf aspectual}: verbs expressing the ability of doing an action &  
\textbf{VN} (2): consider{-}29.9, wish{-}62; 
\textbf{FN} (2): purpose, cogitation\\
\hline\noalign{\smallskip}

\textbf{110} 0{-}{-} & 195 (31\%) --  \emph{anbieten (offer), empfehlen (recommend), fordern (demand)} & {\bf future-oriented}: verbs marking AOs as anticipated, planned &   
\textbf{VN} (89): force{-}59, forbid{-}67, wish{-}62, promote{-}102, urge{-}58.1, order{-}60,
admire{-}31.2, order{-}60, promise{-}37.13	; 
\textbf{FN} (43):  request, preventing \\
\hline\noalign{\smallskip}

000 \textbf{11}{-}  & 15 (2\%)  --  \emph{nachfragen (inquire), anfragen (ask)} & {\bf interrogative}: verbs marking AOs as under investigation  &  
\textbf{VN} (3): estimate{-}34.2, inquire{-}37.1.2, order{-}60; 
\textbf{FN} (1): questioning, request \\
\hline\noalign{\smallskip}

111 \textbf{1}{-}{-}  & 122 (19\%)  -- \emph{bedauern (regret), \"{u}berwinden (overcome), danken (thank)} &  {\bf wh-factual}: opinion verbs marking AOs as factual &   
\textbf{VN} (45): transfer{-}mesg{-}37.1.1, wish{-}62, admire{-}31.2, complain{-}37.8,	conjecture{-}29.5, say{-}37.7; 
\textbf{FN} (18): statement, reveal{-}secret\\
\hline\noalign{\smallskip}

\textbf{110} \textbf{1}0{-} & 30 (5\%)  -- \emph{bef\"{u}rworten (approve), verteidigen (defend), loben (praise)}& {\bf future-oriented wh-factual}: opinion  verbs marking AOs as future-oriented and factual & 
\textbf{VN} (15): admire{-}31.2, allow{-}64, transfer{-}mesg{-}37.1.1, suspect{-}81,
characterize{-}29.2	, neglect{-}75, want{-}32.1, defend{-}85,	
comprehend{-}87.2; 
\textbf{FN} (10): judgment, grant{-}permission, defend, experiencer{-}focus, judgment{-}communication,	
justifying,	 hit{-}or{-}miss, statement,	reasoning, tolerating, grasp \\ 
\hline\noalign{\smallskip}

1{-}{-} \textbf{11}{-} & 120 (19\%)  --  \emph{beschreiben (describe), h\"{o}ren (hear), erinnern (remember)} & {\bf wh$/$if -factual}: objective verbs marking AOs as factual & 
\textbf{VN} (55): discover{-}84, say{-}37.7, see{-}30.1, comprehend{-}87.2,
rely{-}70, seem{-}109, consider{-}29.9, transfer{-}mesg{-}37.1.1, estimate{-}34.2, inquire{-}37.1.2; \textbf{FN} (23): perception{-}experience, statement, cogitation, grasp \\
\hline\noalign{\smallskip}

\textbf{110} \textbf{11}{-} & 48 (8\%)  -- \emph{festlegen (determine), absch\"{a}tzen (assess), lehren (teach)} & {\bf future-oriented wh$/$if-factual}: objective verbs marking AOs as future-oriented and factual &  \textbf{VN} (28): estimate{-}34.2, rely{-}70, indicate{-}78, transfer{-}mesg{-}37.1.1, correspond{-}36.1, conjecture{-}29.5,	discover{-}84,	 say{-}37.7; 
\textbf{FN} (16): predicting, education{-}teaching, assessing, reliance, reasoning \\
\hline\noalign{\smallskip}

111 0{-}{-} & 66 (10\%)  --  \emph{vorwerfen (accuse), bestreiten (deny), f\"{u}rchten (fear)} & {\bf non-factual}: verbs marking AOs as not resolvable re. their factuality &  
\textbf{VN} (28): conjecture{-}29.5,	 wish{-}62, complain{-}37.8, admire{-}31.2; 
\textbf{FN} (13): statement, reveal{-}secret, experiencer{-}focus, certainty \\
\noalign{\smallskip}\hline
  \end{tabular}
  \caption{The 632 verbs in 8 semantic classes (5 verbs show idiosyncratic behavior). Signature substrings in bold correspond to meaning components, which (along with tentative class names) are based on Sec.~\ref{hypo}. The cross-lingual semantic characterization shows aligned  VerbNet (VN) classes covering 265 (42\%) verbs and aligned FrameNet (FN) frames covering 126 (20\%) verbs, see Sec.~\ref{intrinsic}.\footnote{VerbNet subclass suffixes are omitted for lack of space.}}
  \label{table:classes}
\end{table*}
We employ the following  findings on correspondences 
between verb meaning and syntax in order to infer semantic classes 
 from the syntactic signatures. This gives also rise to tentative names (labels) for the corresponding meaning components.

{\bf Factuals:  the \emph{that-wh} and the \emph{that-wh$/$if} alternation.}
Verbs that are able to alternatively take \emph{that} and \emph{wh}-clauses coerce the embedded
interrogative and declarative clauses into factual AOs, corresponding to a particular fact \cite{ginzburg1996}.
Among the verbs showing the \emph{that-wh} alternation are the well-known factive verbs \cite{kiparsky70}
(e.g.,
\emph{She proves \underline{that} she exists.} vs. \emph{She proves \underline{who} she is.}  vs. \emph{He proves \underline{whether} he can mine gold.}).

There is a further distinction among these verbs regarding the ability to take an embedded
\emph{if$/$whether}-question: \newcite{Schwabe2009} show that the \emph{that-wh$/$if} alternation is connected
to objective verbs entailing the existence of an independent witness, whereas the \emph{that-wh} alternation
(i.e., an \emph{if$/$whether}-question is not possible) occurs with non-objective  verbs
(e.g., \emph{He regrets \underline{whom} he ended up with.}  vs. 
\emph{$\star$He regrets \underline{whether} he ended up playing this game.}).

{\bf ``Aspectuals'':  the inability to take \emph{that}-clauses and \emph{to}-infinitives in the past tense.}
Recently, linguistic research has increasingly addressed particular semantic aspects of \emph{to}-infinitives.
\newcite{kush2011} has investigated AOs that can neither be realized  as \emph{that}-clause nor as \emph{to}-infinitive in the past tense
(e.g., 
\emph{She hesitates to answer.} vs. \emph{$\star$She hesitates to have answered.}\footnote{This is the literal translation of the German equivalent to English. In English, the ing-form in the past would be more typical instead of a \emph{to}-infinitive in the past tense.} vs. \emph{$\star$She hesitates that ...})
These AOs are selected by control verbs\footnote{``Control'' refers to the co-reference between the
implicit subject of the infinitival argument and syntactic arguments
in the main clause, either the subject (subject control) or direct object (object control).} and  can be characterized as mental actions. \newcite{kush2011} points out 
that the verbs selecting those AOs have an aspectual meaning in common.

 {\bf Future orientation: \emph{to}-infinitives in the present tense and the inability to take \emph{to}-infinitives in the past tense.}
\newcite{laca:hal} has investigated verbs across English and Spanish that embed future-oriented AOs. 
 Only future-oriented AOs can be used with
 future-oriented adverbials, such as \emph{tomorrow}, and these AOs are often realized as non-finite constructions,
e.g.,  \emph{to}-infinitives. She points out that not only 
 control verbs take future-oriented AOs, but also verbs expressing attitudes of preference. 
This finding implies that such future-oriented AOs are typically incompatible with past-oriented adverbials
(e.g.,  \emph{yesterday}) and verb forms in the past tense (e.g.,
\emph{$\star$She plans having finished the assignment yesterday.}).

\subsection{Mapping to Meaning Components}
We automatically infer semantic classes based on a manually constructed mapping between the syntactic signatures from \newcite{eckle99}  and the meaning components grounded in syntax summarized in Section \ref{hypo}.\footnote{We did not consider verbs can  be used with all kinds
of clausal and infinitival arguments, such as the majority of communication verbs (e.g., \emph{comment, whisper}).}

We constructed this mapping in two steps:
In a first step, the signatures are aligned to the meaning components from Section \ref{hypo} based on substrings of the signatures: future-orientation matches the 110 prefix,
aspectual the 010 prefix, and factuality matches 1's in fourth or fifth position.
It is important to point out that future-orientation can be combined with factuality: this corresponds to an independent matching of the 110 prefix and the factuality substring. While this combination may seem contradictory, it reflects the lexical data and shows that also weak forms of factuality (``it will most likely be factual at some point in the future") are expressed in language.

In a second step,
the pre-aligned signatures are merged, if the remaining slots of the signature are either 1 or 0 (i.e. the respective argument types can or can not occur); in the resulting merged signature, these slots are left underspecified. Merging the signatures in this way yields 8 partially underspecified signatures which correspond to the final semantic classes. 
This procedure covers more than 99\% of the 637 verbs under investigation: only 5 verbs showed idiosyncratic syntactic behavior, 4 of those are verbs that can take an AO as subject (e.g., {\em bedeuten (mean)}). As a consequence of the automatic part of this procedure, every verb is assigned to exactly one class -- a simplification which we plan to resolve as part of future work.

Table \ref{table:classes} provides an overview and a characterization of these classes, also showing the final signatures and their substrings which correspond to the meaning components. The non-factual class is derived from the wh-factual class: the only difference is the inability to take
a \emph{wh}-clause (e.g. \emph{$\star$ He hopes, \underline{when} he will succeed.}).

While the descriptions of the meaning components and the class names are inspired from research in linguistics (typically a very deep analysis of only few verbs), transferring them  to our verb resource -- which is of much larger scale -- inevitably leads to outlier verbs in the classes, e.g., verbs that do not strictly match the class label. Examples include verbs such as {\em \"{u}berlegen (consider)} in the \emph{wh$/$if-factual} class (not covering the future-oriented meaning component) or {\em schaden (harm)} as \emph{non-factual rather} than as \emph{wh-factual}.
For this reason, and also because of the assignment of highly polysemous verbs to only one class, the definitions of meaning components and the class names should rather be considered as loose, providing a first tentative semantic characterization of the modality classes.

In sum, this section presented an inventory of modal meaning components that we primarily synthesized from research in linguistics. The classification work is strictly grounded in syntactic properties of the verbs and was not targeted a priori at modal meanings.

\section{Evaluation}
\subsection{Linking to Semantic Classes} \label{intrinsic}
Our first set of experiments aims at refining the initial semantic characterization of the classes
by linking them to  independently constructed semantic classifications \emph{at the word sense level}.
Specifically, we consider three different semantic classifications from computational lexicons, which have been created by  linguistic experts: (i) the so-called semantic fields in GermaNet, grouping verb senses into
15 coarse classes, such as \emph{perception, emotion},
(ii) the verb classes given in VerbNet, and (iii)
the Frame-semantic frames in  FrameNet.
As the GermaNet and FrameNet classes are based on different lexicographic and linguistic theories, we expect an additional semantic characterization from the linking.
The VerbNet classes, which also follow Levin's hypotheses, however, are used to investigate if the 
syntax-semantics correspondence
is maintained across the related languages German and English.

For this linking experiment, 
we used the UBY framework \cite{uby}\footnote{\url{http://www.ukp.tu-darmstadt.de/uby/}},
containing standardized versions of the above lexicons, as well as a linking between VerbNet and FrameNet on the word sense level.

{\bf Approach} In order to link our classes to verb senses in GermaNet and VerbNet, we developed an automatic linking method based on subcat frame similarity. Recognizing subcat frame similarity requires
 a common standardized format for the otherwise incomparable frames.
 UBY provides such a standardized format which has been presented in detail by \newcite{TUD-CS-2012-0024}. It represents subcat frames  uniformly across German and English, and
at a fine-grained level of individual syntactic arguments.
Our linking approach is based on the following hypothesis:
Two verb senses with equivalent lemmas are equivalent, if they have similar subcat frames.\footnote{This  approach is applicable for GermaNet, because GermaNet contains fine-grained syntactic subcat frames.} Our  method interprets the pairs of verb and subcat frame listed in our classification\footnote{We consider
only verb senses that are compatible with AOs, as indicated by subcat frames with
clausal or non-finite arguments.} as senses.
While we do not claim that this hypothesis is sufficient in general, i.e., for all verb senses,
we found that it is valid for the subset of senses belonging to the class of AO-selecting verbs.

The cross-lingual linking of our classes to VerbNet senses requires an additional translation step, which we describe first.

{\bf Manual Translation}
While UBY also provides translations between German and English verb senses, e.g., as part of the
Interlingual Index from EuroWordnet (ILI), we found that many of the translations were not present in our target lexicon VerbNet. 
Therefore, the main author of this paper, a native speaker of German with a good proficiency in English, 
translated the AO-compatible verbs (i.e., word senses) manually using Linguee\footnote{Linguee (\url{http://www.linguee.de/}) is a translation tool combining an editorial dictionary and a search engine processing bilingual texts. In particular, it provides a large variety of contextual translation examples.} and dictionaries. This took about 7 hours.

For 23 German verbs, we could not find any  equivalent 
lexicalized translation, because these verbs express very fine-grained semantic nuances.
For example, we did not find an equivalent English verb for a few verbs in the aspectual class
but only a translation
consisting of an adjective in combination with \emph{to be}.
Examples include \emph{be easy (leichtfallen)}, \emph{be willing (sich bereitfinden)}, \emph{be capable (verm{\"o}gen)}, which have German equivalents that are lexicalized as verbs.
As a result, we arrived at translations for 614 out of 637 German verbs.
These 614 German verbs are translated to  413 English verbs, indicating that the English translation has a more general meaning in many cases.

{\bf Automatic Verb Sense Linking}
Our algorithm links a German verb sense (or its English translation) with a GermaNet (or  VerbNet) sense, 
if the subcat frames of both verb senses
have the same number of  arguments and 
if the arguments have certain features in common.\footnote{We do not link the subcat frames, but we do compare them across the related languages German and English to determine their similarity in the context of linking.}
For example,
to create a link to GermaNet, features such as the complementizer of clausal arguments and the case of noun phrase arguments have to agree. In a similar way, the linking to VerbNet  is based on a comparison of  German subcat frames and English subcat frames -- which are represented uniformly across German and English.
In Section \ref{sl}, we provide more details about the algorithm.

{\bf Results}
According to a manual evaluation of a random sample of 200 sense pairs, the automatic verb sense linking yielded 
an accuracy of $89.95\%$ for the linking to GermaNet,
and $87.54\%$ for the linking to VerbNet ($\kappa$ agreement on the sample  annotated by two annotators was $0.7$ and $0.8$, respectively). The main types of errors in the linking to GermaNet and VerbNet are due to  specific syntactic features of the subcat frames which diverge and are not considered in the automatic linking. 
The differences regarding these specific features are due to cross-lingual differences (VerbNet, e.g., verb phrase arguments with ing-form) and diverging linguistic analyses of particular constructions (GermaNet, e.g., constructions with \emph{es (it)}), see also \newcite{TUD-CS-2012-0024}.

By linking the verbs in our classification to semantic classes in GermaNet, VerbNet and FrameNet, we obtain a three-way semantic characterization of our  classes.
The linking to the GermaNet semantic fields covers 270 (43\%) of the source verbs. Of these,
219 (81\%) are linked to the three semantic fields \emph{cognition}, \emph{communication} and \emph{social}. Fewer verbs (32 (12\%)) are linked to the semantic fields \emph{emotion, perception, change}. Semantic fields not among the target classes are \emph{consumption, competition, contact, body}  and \emph{weather}.

Table \ref{table:classes} summarizes the linking to VerbNet and FrameNet and shows how many verbs from each source class could be  linked to any of the classes in VerbNet or FrameNet.\footnote{Based on the percentage of source class members linked to any of the target classes, we only display target classes with an overlap of at least  $1.8\%$ due to space constraints.} As the class distribution of the verb subsets covered by our linking-based evaluation is similar as for the original classes, we consider our evaluation as valid, although less than 50\% of all verbs could be evaluated this way. 

The target classes in
VerbNet and FrameNet 
reveal   meaning components that are on the one hand unique for individual classes, and
on the other hand  shared across several German classes. 

The \emph{future-oriented} class contains object control verbs 
(e.g., force{-}59, forbid{-}67
in VerbNet, and  \emph{request, preventing} in FrameNet).
 The \emph{wh$/$if-factual} class is unique regarding the \emph{cognition and perception} verbs 
(e.g., discover{-}84, see{-}30.1{-}1, and \emph{perception-experience}).
The \emph{future-directed wh$/$if-factual} class also contains \emph{objective assessment}  verbs,
as shown by the estimate{-}34.2 class.
The verbs in the two \emph{wh-factual} classes share meaning components as well, 
 as shown by the \emph{opinion verb} classes admire{-}31.2 and defend{-}85 in VerbNet or
\emph{judgment, tolerating} in FrameNet.

While there are also other VerbNet and FrameNet classes shared across several classes,
they turned out to be very general and underspecified regarding their meaning,
thus not contributing to a more fine-grained semantic characterization.
For example, the
 conjecture{-}29.5 class assembles quite diverse conjecture verbs, e.g. 
verbs expressing opinion (\emph{feel, trust}) and factuality (\emph{observe, discover}).
A similar observation holds for the \emph{statement}  frame in FrameNet.

\subsection{Analysis of Frequency and Polysemy}
\begin{table}[t]
\centering
\footnotesize
\begin{tabular}{@{}p{2.2cm}|ll|l||l}
\hline\noalign{\smallskip}
\bf Verb class &  \bf Wiki &   \bf Web  &  \bf News & \bf News Eng. \\
\noalign{\smallskip}\hline\noalign{\smallskip}
all &  25.85 &  50.58 & 33.91 & 25.31 \\
\hline\noalign{\smallskip}
aspectual  & 0.90 & 0.80 & 1.44 & 1.96\\
future-oriented & 9.45 & 23.04 & 13.65 & 12.58 \\
interrogative & 0.01 & 0.05 & 0.05 & 0.65\\
wh-factual & 4.26 & 17.89 & 4.99 & 3.48 \\
fo. wh-factual & 0.29 & 0.28 & 0.85 & 1.14\\
wh$/$if -factual & 3.02 & 2.54 & 3.53 &  5.20\\
fo. wh$/$if-factual & 2.36 & 1.77 & 3.14 & 5.75\\
non-factual & 4.29 & 3.36 & 4.84 & 3.57\\

\noalign{\smallskip}\hline
  \end{tabular}
  \caption{Percentage of classes in corpora:  German Wikipedia (Wiki), SDeWaC (Web), Tiger (News); 
 English Reuters corpus (News Eng.).}
  \label{table:distribution}
\end{table}
In order to assess the usefulness of the verb resource for NLP tasks, we determined
the lemma frequency of all verbs in the 8 classes 
 in SDeWaC \cite{Faass2013}, a cleaned version of
the German DeWaC corpus \cite{Baroni:2006}. 
A ranking of the verbs according to their lemma frequency showed that 89\% of the verbs occur more than 50 times in SDeWaC.\footnote{In the verb resource we provide for download, we included this frequency information in order to enable frequency-based filtering.}

We also analyzed the frequency distribution of the 8 verb classes in two other German corpora belonging to different genres, and
also for English, see
Table \ref{table:distribution}:\footnote{Details of the computation of the verb lemma frequency lists are given in the appendix \ref{freq}.}
 encyclopedic text (the German Wikipedia\footnote{\url{www.wikipedia.de}, dump of 2009-06-18}),
German  newspaper text 
(the Tiger corpus \cite{tiger20014}),
and the English Reuters-21578 corpus.\footnote{Reuters-21578, Distribution 1.0,
see \url{http://kdd.ics.uci.edu/databases/reuters21578/reuters21578.html}.}
Table \ref{table:distribution} shows that the large verb classes constitute a substantial proportion of verb occurrences
 across different genres.
This suggests that the verb classes might be useful features for various text classification tasks.

We performed a further analysis of the polysemy of the German and English verbs in our classes relative to several fine and coarse word sense inventories. Regarding GermaNet, there are 2.28 senses per verb (1.53 for all GermaNet verbs), whereas WordNet lists 5.11 senses per verb (2.17 for all WordNet verbs). In VerbNet, we find 1.74 senses per verb (1.42 for all VerbNet verbs), and in FrameNet 1.96 (1.52 for all FrameNet verbs). This analysis shows that the task of automatic sense linking is particularly hard for the category of AO-selecting verbs we consider. Whether the polysemy is an issue for any application where the verb classes are used as features is not a priori clear and depends on the task at hand.

\subsection{Textual Entailment Experiment}  \label{extrinsic}
For an extrinsic evaluation, we  investigated the usefulness of the German \emph{and} the English verb classes as features in recognizing textual entailment (RTE). In RTE, the task is to determine
whether for a pair of text fragments -- the text T and the hypothesis H -- the meaning of H is entailed by T
\cite{rte2006}; for non-entailing pairs, sometimes a further category ``unknown" is used as a label. 

We employed a simple classification-based approach to RTE and trained and evaluated a Naive Bayes classifier
on the
test sets of three RTE benchmarks, using 10-fold cross validation: the English RTE-3 data \cite{rte07}  and their
German translation\footnote{\url{http://www.dfki.de/~neumann/resources/RTE3_DE_V1.2_2013-12-02.zip}} (the development sets and the test sets each consist of 800 pairs), and an expanded
version of the English RTE-3 data from the Sagan Textual Entailment Test Suite 
\cite{saganSuite} consisting of 2974 pairs. While the German dataset provides a two-way classification of the T-H pairs, the two English datasets provide a three-way classification, also using the ``unknown" label.
We used the DKPro TC framework \cite{dkprotc}  for
classification and applied POS tagging and lemmatization as preprocessing.

As a baseline feature, we use the word overlap measure 
between two T-H pairs (no stopword filtering, no lemmatization, no normalization of overlap score), which is quite competitive on the RTE-3 data, because this dataset shows a high
difference in word overlap between positive (entailment) and negative (no entailment) pairs \cite{rte09}.

An analysis of the development set of the German RTE-3 data showed that 62\% of the pairs contain at least one occurrence of any of the verbs from the classification in either T or H. However, 
T and H fragments display no statistically significant differences\footnote{All significance scores in this paper are based on Fisher's exact test at significance level p$<$0.05.} regarding the occurrences of any of the verb classes. 

A detailed analysis revealed that pairs without entailment are often characterized by a mismatch between T and H regarding the presence of factuality markers. For example, the presence of verbs indicating uncertainty (all classes apart from 
wh-factual and wh$/$if-factual) in T and an absence of such verbs in H might indicate non-entailment 
as in the following not entailing pair from the English RTE3 development set where ``long" signals non-factuality, but ``researching" signals factuality: 
\vspace{-0.2cm}
\begin{myquote}
T: \emph{The BBC's Americas editor Will Grant says many Mexicans are tired of conflict and long for a return to normality.}\\
H: \emph{Will Grant is researching a conflict with Mexicans.}
\end{myquote}
\vspace{-0.2cm}
Thus, an insufficient overlap of modality markers in T and H might actually indicate non-entailment, but lead to an incorrect classification as entailment when considering only word overlap.

Accordingly, we implemented a factuality-mismatch feature both for German and for English, based on our new German and English classes. This feature is  similar to the word overlap feature but with lemmatization and normalization of overlap score.
Verb class counts are based on verb lemma counts of the member verbs; for English verbs that are  members of more than one class, we included all verb classes in our factuality-mismatch feature.\footnote{In the German part, every verb is assigned to one class, while the translation to English resulted in  22\% of the English verbs being members in more than one class. However, only 11\% of the multiple class assignments involve a combination of factual and uncertainty classes.}
Table \ref{entail} shows the results.
While the differences for RTE-3 DE and
RTE-3 EN are not statistically significant, the factuality-mismatch feature yielded a small but significant improvement
on the expanded RTE-3 EN  dataset. This is due to the different nature of the
 expanded RTE dataset, which was created using a paraphrasing technique. As a result, the number of occurrences of verbs from our classes increased, and the factuality-mismatch became a discriminative feature for distinguishing between CONTRADICTION and UNKNOWN/ENTAILMENT.

Considering the fact that we employed only simple overlap features that do not rely on dependency parsing and did not perform any word sense disambiguation,
these  results suggest that the verb classes might be promising features for RTE, both for German
\emph{and} English. As factuality can be expressed by a variety of further linguistic means, including modal verbs, negation, tense and certain adverbs, investigating the combination of our verb classes with other modality signals might be especially promising as part of future work.

\begin{table}[h]
\footnotesize
\begin{center}
\begin{tabular}{@{}llll}
\hline\noalign{\smallskip}
\bf    & \bf RTE-3 DE & \bf RTE-3 EN & \bf RTE-3 EN exp. \\ 
\noalign{\smallskip}\hline\noalign{\smallskip}
WO  & 59.87 & 54.75 & 54.98 \\
WO$+$FM & 59.25 & 54.62 & \bf 58.81 \\
\noalign{\smallskip}\hline
\end{tabular}
\end{center}
\caption{Accuracy of a Naive Bayes classifier (10-fold cross validation on
the test sets) with word overlap (WO) and additional
factuality-mismatch (WO$+$FM) features.}
\label{entail}
\end{table}

\section{Results and Discussion}
Our construction of semantic classes from the syntactic behavior of AO-selecting verbs results in an inventory of modal meanings that emerged from a large lexical resource. 
The main result of the linking based evaluation is a detailed semantic characterization of the inferred classes -- a prerequisite for using them in NLP tasks in an informed way.
 The semantic classes seem to be particular suited for tasks related to opinion analysis, textual inference,
 or argumentation mining.
In this context, the relationship between our large resource of lexical verbs and the closed class of modal verbs might be an interesting question for future research. 
 
Most of all, the linking to GermaNet and FrameNet shows that it is indeed possible to narrow down meaning components for Levin classes.
Moreover, the results of the linking to VerbNet also provide support for Levin's 
hypothesis that the 
correspondences between verb syntax and meaning described for English largely apply to the related language 
German as well \cite{levin2015cross}.

The English version of the semantic classes which we created by means of translation has
the same  semantic properties as the German classes.
However, the syntactic properties of the English classes are not fully specified, because English has additional kinds of non-finite arguments, such as ing-forms or bare infinitives. Therefore, it might be interesting to address this question in the future and to build a similar semantic classification for English from scratch, in particular in the context of extracting modality classes from corpora.
This would require an adaptation of the syntactic signatures, considering the various kinds of non-finite arguments particular to English.
Based on large subcategorization lexicons available for English (e.g. COMLEX \cite{DBLP:conf/coling/GrishmanMM94} or VerbNet), it should be feasible to derive such signatures and to construct a mapping of signatures to modality aspects in a similar way as for German.

 The question whether the syntactic signatures can be recovered in large corpora is particularly interesting, 
 because this would allow extending the existing classes and to also acquire AO-selecting adjectives and nouns. We plan to investigate this question as part of future work.

\section{Conclusion}
We  inferred semantic classes from a large syntactic classification of German AO-selecting verbs 
based on findings  from formal semantics about
correspondences between verb syntax and meaning. 
Our thorough evaluation and analysis yields detailed insights into the semantic characteristics of the inferred classes, and we hope that this allows an informed  use of the resulting resource in various semantic NLP tasks.

\section*{Acknowledgments}
This work has been supported by the Volks\-wagen
Foundation as part of the Lichtenberg-Professorship Program under grant No. I/82806 and by the German Research  Foundation  under  grant  No. GU 798/17-1
and No. GRK 1994/1.
We thank the anonymous reviewers for their valuable comments. 
Additional thanks go to  Anette Frank, Iryna Gurevych and Ani Nenkova for their helpful feedback on earlier versions of this work.


\appendix

\section{Supplemental Material}
\label{sec:supplemental}
\subsection{Verb Lemma Frequency List} \label{freq}
In order to count the occurrences of verb lemmas in the German corpus SDeWaC, we used a reader and pre-processing components (i.e., the LanguageTool segmenter and the TreeTagger for POS tagging and lemmatization)
from the DKPro Core collection \cite{TUD-CS-2014-0864}. From DKPro Core, we also used a component that detects separated particles of German particle verbs and replaces the lemma of the verb base form annotated by the TreeTagger by the true lemma of the particle verb.
Our verb lemma counting pipeline is available at \url{github.com/UKPLab/acl2016-modality-verbclasses}.

\subsection{Verb Sense Linking} \label{sl}
\begin{table}[h]
\footnotesize
\begin{tabular}{@{}l}
\toprule
Sense linking based on subcategorization frames\\
\midrule
get lexical entry $le_s$ of source verb $v_s$\\
get equivalent verb $v_t$ in target lexicon\\
get lexical entry $le_t$ of target verb $v_t$ \\

{\bf forall} frame $f_i$ in $le_s$\\
\hspace{0.2cm} get listOfArguments $l_i$ of $f_i$ \\
\hspace{0.2cm} {\bf forall} frame $f_j$ in $le_t$\\
\hspace{0.4cm} get sense $s_j$ of frame $f_j$ \\
\hspace{0.4cm} get listOfArguments $l_j$ of $f_j$\\
\hspace{0.4cm} {\bf if} $size(l_i$) = $size(l_j$)\\
\hspace{0.4cm} AND $features(l_i$) = $features(l_j$)\\
\hspace{0.6cm} link $(v_s,f_i)$ and $s_j$\\
\hspace{0.4cm} {\bf end if}\\
\hspace{0.2cm} {\bf end for}\\
{\bf end for}\\
\bottomrule
\end{tabular}
\caption{Algorithm for verb sense linking.}
\label{table:labelling}
\end{table}
For the linking-based evaluation, we used UBY (version 0.7.0) versions of
the following three resources:
 the German wordnet GermaNet (version 9.0),
the English lexicons VerbNet (version 3.2) and FrameNet 
(version 1.5).

The algorithm for cross-lingual verb sense linking is given in pseudo-code in Table \ref{table:labelling}.
The implementation is available at \url{github.com/UKPLab/acl2016-modality-verbclasses}.

\end{document}